\documentclass{article}

    \PassOptionsToPackage{numbers, compress}{natbib}

\usepackage[final]{neurips_2023}

\usepackage[utf8]{inputenc} 
\usepackage[T1]{fontenc}    
\usepackage{hyperref}       
\usepackage{url}            
\usepackage{booktabs}       
\usepackage{amsfonts}       
\usepackage{nicefrac}       
\usepackage{microtype}      
\usepackage{xcolor}         
\usepackage{graphicx}
\usepackage{amsmath}
\usepackage{multirow} 
\DeclareMathOperator*{\argmax}{arg\,max}

\definecolor{mypink}{RGB}{239,43,159}

\def\Var{{\textrm{Var}}}
\def\E{{\textrm{E}}}

\title{LLaVA-UHD: an LMM Perceiving Any Aspect Ratio and High-Resolution Images}

\author{%
    Ruyi Xu\,$^{1}$  \quad Yuan Yao\,$^2$\thanks{Corresponding Authors} \quad Zonghao Guo\,$^{3}$ \quad Junbo Cui\,$^{1}$ \quad Zanlin Ni\,$^1$ \quad Chunjiang Ge\,$^1$ \\
    \quad 
    \textbf{Tat-Seng Chua}\,$^2$ \quad \textbf{Zhiyuan Liu}\,$^1$ \quad \textbf{Maosong Sun\,$^1$} \quad \textbf{Gao Huang}\,$^{1*}$ \\
    $^1$Tsinghua University \quad
    $^2$National University of Singapore \\ 
    $^3$University of Chinese Academy of Sciences\\
    \texttt{xrorrim@gmail.com} \quad
    \texttt{yaoyuanthu@gmail.com} \quad
    \texttt{gaohuang@tsinghua.edu.cn} \quad
}

\begin{document}

\maketitle

\vspace{-10mm}
\begin{center}
    \large{\color{mypink} \url{https://github.com/thunlp/LLaVA-UHD}}
\end{center}

\begin{abstract}
  Visual encoding constitutes the basis of large multimodal models (LMMs) in understanding the visual world. Conventional LMMs process images in fixed sizes and limited resolutions, while recent explorations in this direction are limited in adaptivity, efficiency, and even correctness. In this work, we first take GPT-4V and LLaVA-1.5 as representative examples and expose systematic flaws rooted in their visual encoding strategy. To address the challenges, we present LLaVA-UHD, a large multimodal model that can efficiently perceive images in any aspect ratio and high resolution. LLaVA-UHD includes three key components: (1) An image modularization strategy that divides native-resolution images into smaller variable-sized slices for efficient and extensible encoding, (2) a compression module that further condenses image tokens from visual encoders, and (3) a spatial schema to organize slice tokens for LLMs. Comprehensive experiments show that LLaVA-UHD outperforms established LMMs trained with 2-3 orders of magnitude more data on 9 benchmarks. Notably, our model built on LLaVA-1.5\hspace{0.5mm}$_{336\times336}$ supports 6 times larger (i.e., 672$\times$1088) resolution images using only 94\% inference computation, and achieves 6.4 accuracy improvement on TextVQA. Moreover, the model can be efficiently trained in academic settings, within 23 hours on 8 A100 GPUs (vs. 26 hours of LLaVA-1.5).
\end{abstract}

\section{Introduction}

Recent progress in Large Multimodal Models (LMMs)~\citep{2023llava1.6,instructblip2023, li2023monkey,liu2024llava,bai2023qwen} has witnessed a significant surge in vision-language understanding, reasoning, and interaction capabilities. This is achieved by projecting visual signals into Large Language Models (LLMs) to enable their visual perception of the world, where visual encoding strategy plays a fundamental role~\citep{li2023blip2,Alayrac2023Flamingo,liu2024llava}. Real-world images are known to reside in a wide range of aspect ratios and resolutions, presenting significant challenges for LMMs in various applications.

However, most existing LMMs~\citep{chen2023shikra,instructblip2023,liu2024llava} perceive images in a fixed aspect ratio (i.e., 1:1) and a low resolution (i.e., 224$\times$224). The compromise to this simplified setting typically leads to severe shape distortion and blur of image contents. The problem significantly hurts the capabilities of LMMs, especially for fine-grained capabilities, such as small object understanding~\citep{li2023otterhd} and optical character recognition~\citep{ye2023ureader,bai2023qwen,hong2023cogagent}. Moreover, the issue also exacerbates hallucination problems (i.e., producing textual responses not factually grounded in images), since models can only learn to make best guesses to blurred images~\citep{sun2023aligning,yu2023rlhf}.

To achieve image perception in varied aspect ratios and high resolutions for LMMs, there are two main challenges: (1) Adaptivity. Since visual encoders (e.g., CLIP-ViT~\citep{radford2021clip}) are pretrained in fixed resolutions, it can be difficult to deal with images in a wide range of aspect ratios and resolutions. Simple image interpolation that deviates far from the pretraining scenarios can result in out-of-distribution issues. (2) Efficiency. Directly encoding high-resolution images using vision Transformers~\cite{dosovitskiy2020vit} requires quadratic computation cost with respect to image sizes. In addition, it can be even more costly for LLMs to process the large number of visual tokens from high-resolution images (e.g., 4096 tokens for 896$\times$896 images in ViT-L/14).

Moreover, careless visual encoding strategies can even result in systematic flaws in correctness. For example, despite its powerful capabilities in various aspects, it has been commonly reported that GPT-4V~\citep{achiam2023gpt4} can surprisingly struggle in some basic capabilities, such as identifying the number of objects~\citep{yang2023dawn}. The mechanistic cause for such embarrassment remains largely unknown. In this work, we perform the first mechanistic investigation of GPT-4V flaws from the perspective of visual encoding strategy. Our controlled experiments in probing GPT-4V show that the problem can be partially rooted in its visual encoding strategy in dealing with high-resolution images. Investigation on LLaVA-1.5~\citep{liu2023llava1.5}, a representative open-source LMM also shows systematic issues in correctness, indicating their potential vulnerability for adversarial attacks.

To address the challenges, we present LLaVA-UHD, a large multimodal model that efficiently perceives any aspect ratio and high-resolution images. The model has three key components: (1) At the core of LLaVA-UHD is an image modularization strategy that divides native-resolution images into smaller variable-sized slices for efficient and extensible encoding. In comparison to recent works that fit images into several fixed aspect ratios and resolutions~\citep{SPHINX2023,li2023monkey}, the variable-sized slices in LLaVA-UHD enable full adaptivity to native-resolution images without padding or shape-distorting resizing. This is in analogy to the better adaptivity of using water drops vs. ice cubes in full-filling variable-sized glasses. We also show that the strategy guarantees minor deviation from the pretraining setting of visual encoders to maximally retain their capabilities. (2) The visual tokens are condensed by a compression layer to modest lengths, largely reducing the computation for LLMs. (3) Finally, the compressed slice tokens are organized in a spatial schema to inform LLMs about the slice positions in the image.

Comprehensive experiments on 9 benchmarks show that LLaVA-UHD significantly improves the capabilities of LMMs, outperforming established counterparts trained with 2-3 orders of magnitude more data. Notably, our model built on LLaVA-1.5\hspace{0.5mm}$_{336\times336}$ supports 672$\times$1088 resolution images using only 94\% inference computation, and achieves 6.4 accuracy improvement on TextVQA and 3.2 accuracy improvement on POPE. The advantage enlarges with more extreme aspect ratios. We also show that instruction tuning on ViT parameters is sufficient for adaptation to a broad range of images. Moreover, the model can be efficiently trained in academic settings, within 23 hours (vs. 26 hours of LLaVA-1.5) on 8 A100 GPUs.


The contribution of this work can be summarized as threefold: (1) We perform the first mechanistic investigation of GPT-4V from the perspective of visual encoding strategy and expose systematic flaws. (2) We present LLaVA-UHD, a large multimodal model that can efficiently perceive any aspect ratio and high-resolution images. (3) We conduct comprehensive experiments to demonstrate the effectiveness of LLaVA-UHD on 9 popular benchmarks, and also provide analysis for deeper understanding of the model.


%




\section{Pilot Experiments}
\label{sec:pilot_exp}

We start with a pilot experiment on the visual encoding strategies of existing LMMs, taking GPT-4V~\citep{achiam2023gpt4} and LLaVA-1.5~\citep{liu2023llava1.5} as representative examples. GPT-4V is a powerful and most recognized proprietary LMM, while LLaVA-1.5 is one of the most influential open-source LMMs. Despite their strong performance in many aspects, it has been commonly reported that dilemmas can be encountered in some basic capabilities~\citep{yang2023dawn}. For example, GPT-4V is prone to miscounting the object numbers in images, whereas the causes remain largely unknown. 

In this work, we perform the first mechanistic investigation of GPT-4V flaws from the perspective of visual encoding strategy. The key idea is that by using synthetic images as continuous probes, we can evaluate the behaviors of GPT-4V in a highly controlled manner, thereby identifying the underlying causes. Our experimental results indicate that, some systematic flaws of GPT-4V are likely to be rooted in its visual encoding strategy, which can be potentially exploited for adversarial attacks.

\subsection{GPT-4V Experiments}

\smallskip
\textbf{Preliminary.} According to the publicly available information from OpenAI,\footnote{~\url{https://platform.openai.com/docs/guides/vision}} GPT-4V employs two image processing modes: low resolution and high resolution. (1) In low-resolution mode, for an original image with dimensions W and H, the model processes only a low-resolution overview image. (2) In high-resolution mode, besides the overview image, GPT-4V processes additional slices of the original high-resolution image, where each slice has $512\times512$ resolution, resulting in $\lceil \frac{W}{512} \rceil \times \lceil \frac{H}{512} \rceil$ slices in total. In our experiments on GPT-4V's new high-resolution mode, interesting error patterns are observed, prompting an exploration into GPT-4V's underlying visual encoding logic.

\begin{figure*}[t]
\centering
\includegraphics[width=1.0\linewidth]{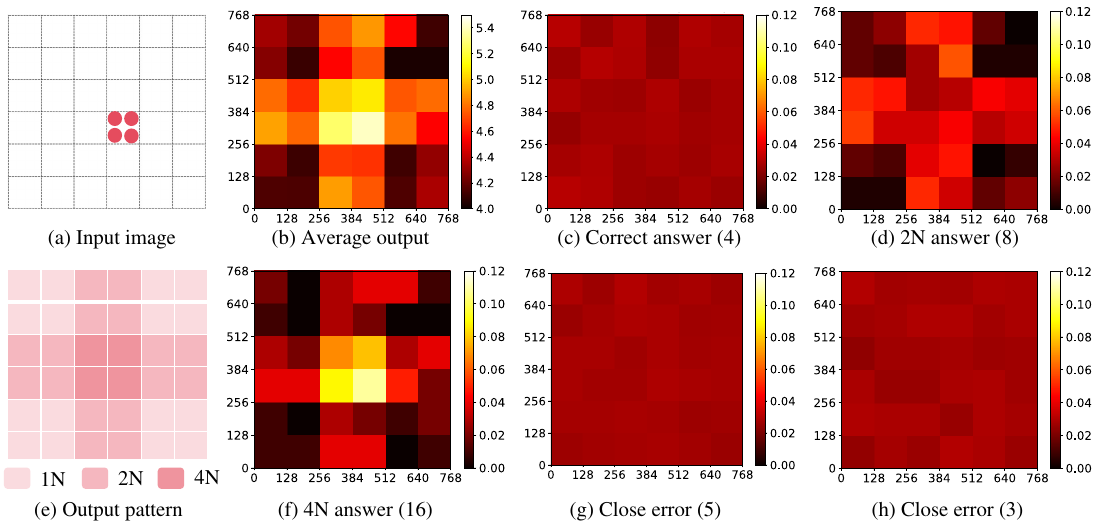}
\caption{Experimental results of GPT-4V in identifying numbers of objects. Note that the dashed lines in (a) are for illustration purposes only, and not presented to GPT-4V.}
\label{fig:gpt4v_exp1}
\end{figure*}

\smallskip
\textbf{How do positions in images influence GPT-4V's behavior?} Our experiments start with a simple instance: Given the image as shown in Fig.~\ref{fig:gpt4v_exp1}(a), we ask GPT-4V: ``How many circles are there in the image?'' We synthesize a series of image variants by changing the positions of circles in the image, and keep the text prompt unchanged. For better reliability, we also synthesize images using other colors and shapes as well, in $\{\text{red}, \text{green}, \text{white}\} \times\{ \text{circle}, \text{triangle}, \text{square}\}$. For each instance, we query 15 times to better approximate the true response distribution.

We calculate the average number answered by GPT-4V for each position in the image, and report the heatmap in Fig.~\ref{fig:gpt4v_exp1}(b). We can observe that the result is highly correlated with object positions in images. Specifically, the patterns are split by $256\times256$ squares, and three interesting patterns can be identified: (1) The central square exhibits the highest response number, (2) the middle edges show a lower number, and (3) the corners are the closest to ground truth.

To investigate the cause, we further separate the model responses by number, and report the distribution across positions for each response in Fig.~\ref{fig:gpt4v_exp1}(c), (d), (f), (g) and (h). Interestingly, besides the correct answers (4: 66.1\%) and close answers (5: 16.6\%, 3: 10.2\%), it turns out that the remaining two abnormal answers (8: 5.2\%, 16: 1.9\%), which doubles and quadruples the ground truth, account for the error pattern in Fig.~\ref{fig:gpt4v_exp1}(b). Combining the results with the public information from OpenAI, we hypothesize the most likely cause is that, there are overlaps in the slices of GPT-4V when the image resolution is not divisible by 512.\footnote{Note that the issue is different from the overlapping sliding windows in CNNs, since the overlaps in GPT-4V is inconsistent across different resolution images.} As illustrated in Fig.~\ref{fig:gpt4v_exp1}(e), the overlapping areas between two slices will double the number, and the overlapping areas between four slices will quadruple the number.\footnote{Note that besides visual encoding strategies, model behaviors are also influenced by the accumulated training dynamics and RLHF. Therefore the double/quadruple effect does not dominate the results. All results are from GPT-4V on 03-05-2024.}

\begin{figure*}[t]
\centering
\includegraphics[width=1.0\linewidth]{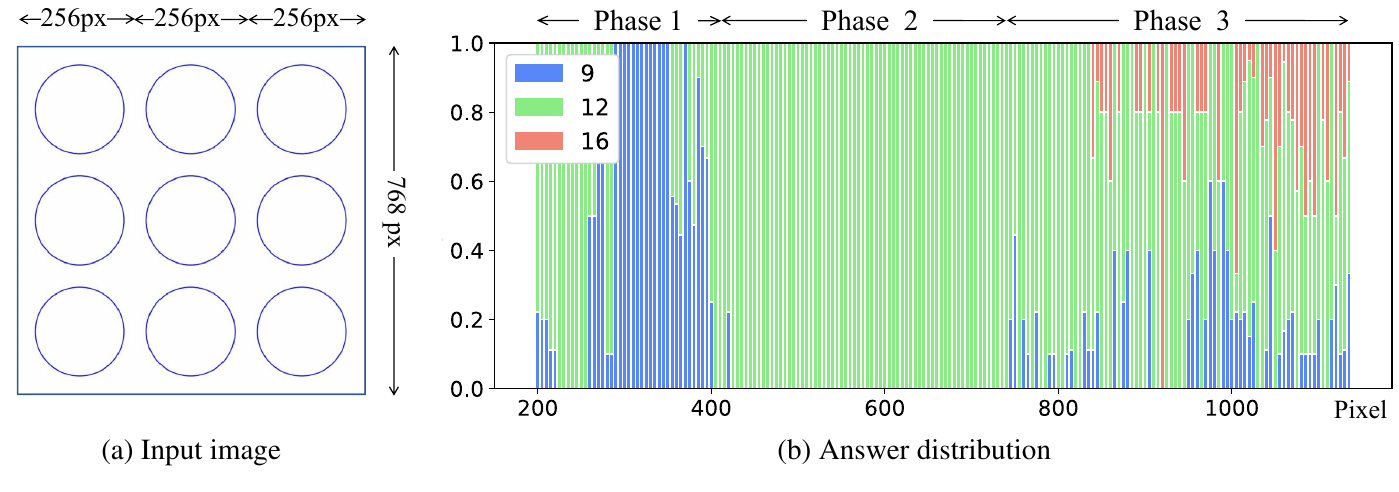}
\caption{Results on probing GPT-4V via continuously changing image resolutions.}
\label{fig:gpt4v_exp2}
\end{figure*}

\smallskip
\textbf{How do image resolutions influence GPT-4V's behavior?} To verify the hypothesis, we further probe GPT-4V through continuously changing image resolutions. Specifically, we proportionally resize the image in Fig.~\ref{fig:gpt4v_exp2}(a) into different resolutions, and query about the object number in the same way. For each resolution, we repeatedly query 30 times for better reliability.

We report the experimental results in Fig.~\ref{fig:gpt4v_exp2}(b). We observe that the model responses show a significant phase change with image resolutions: (1) In phase 1, since there are no image slices, most answers are correct; (2) In phase 2, answer 12 dominates the responses possibly due to the incomplete circles in each slice. (3) Phase 3 shows mixed answers of 9, 12 and 16. Note that 16 can be well explained by the error pattern in Fig.~\ref{fig:gpt4v_exp1}(e). We refer readers to Section~\ref{sec:GPT-4V-illustration} for a more detailed illustration of each phase. Besides, we also notice that many abnormal phenomenons in Fig.~\ref{fig:gpt4v_exp2}(b) cannot be perfectly explained yet, which we leave for future work.

In conclusion, these experimental findings shed light on GPT-4V's potential vulnerabilities in high-resolution image processing, warranting further investigation into the implications of these weaknesses and the development of strategies to counter potential adversarial attacks on LMMs.

\begin{figure*}[t]
\centering
\includegraphics[width=1.0\linewidth]{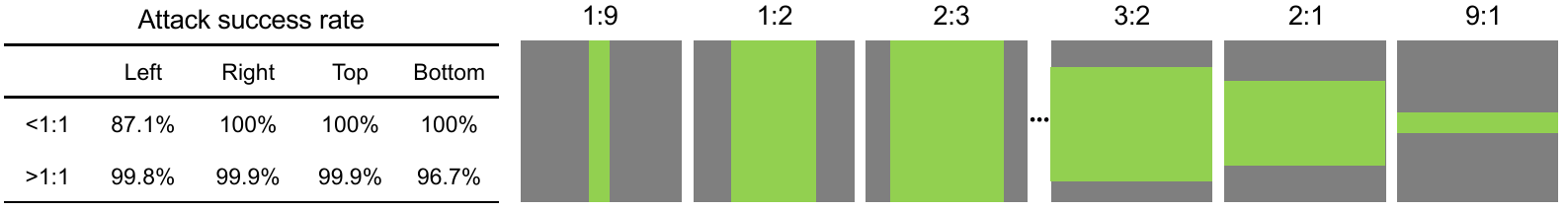}
\caption{Experimental results of adversarially attacking LLaVA-1.5 using input images containing padding pixels. Left: Attack success rates where LLaVA-1.5 ignores the grey area and answers the color of the central rectangle (e.g., green). Right: Synthesized input images containing (1) a rectangle in varied aspect ratios, and (2) padding pixels.}
\label{fig:llava_exp}
\end{figure*}
\vspace{-1em}

\subsection{LLaVA-1.5 Experiments}

To deal with images with varied aspect ratios, LLaVA-1.5 pads the input images into squares before feeding them into the visual encoder. This encoding method results in a waste of computation for non-square images. For example, a 1:4 image has only 25\% effective computation after padding into squares. To quantify the influence, we train an unpadded version of LLaVA-1.5, by fitting the ViT position embedding into the aspect ratio of input images using 2D interpolation. The resultant image tokens remain no more than 576 as in LLaVA-1.5 (see Section~\ref{sec:encoding}). From the experimental results in Table~\ref{tab:module_ablations}, we observe that adaptive aspect ratio encoding without padding consistently improves the performance of LLaVA-1.5.

Another issue of padding is that, the model essentially cannot know whether the padding-like pixels come from image pre-processing or an actual part of the original input image. To demonstrate this issue, we synthesize a series of input images as in Fig.~\ref{fig:llava_exp}(right), where blue/green/red rectangles in various aspect ratios are surrounded by grey (i.e., the color of LLaVA-1.5's padding RGB value). Given the input image, we prompt: ``What is the color of the left/right/top/bottom most area?'' From the results in Fig.~\ref{fig:llava_exp}(left), we observe that LLaVA-1.5 neglects the grey input areas (considering them as padding), and faithfully responds with the color of the central rectangle.





 

\subsection{Conclusions on Pilot Experiments}

In summary, both powerful proprietary LMMs such as GPT-4V and open-source LLaVA-1.5 have systematic issues in their underlying visual encoding strategies. The results show that visual strategies must be designed with caution. Common practices such as padding, shape-distorting resizing, and repetitive slicing can result in a waste of computation, a loss of model capability, and even vulnerability to adversarial attacks. Therefore, there is an urgent need for more adaptive and efficient visual encoding methods.





\begin{figure*}[t]
\centering
\includegraphics[width=1.0\linewidth]{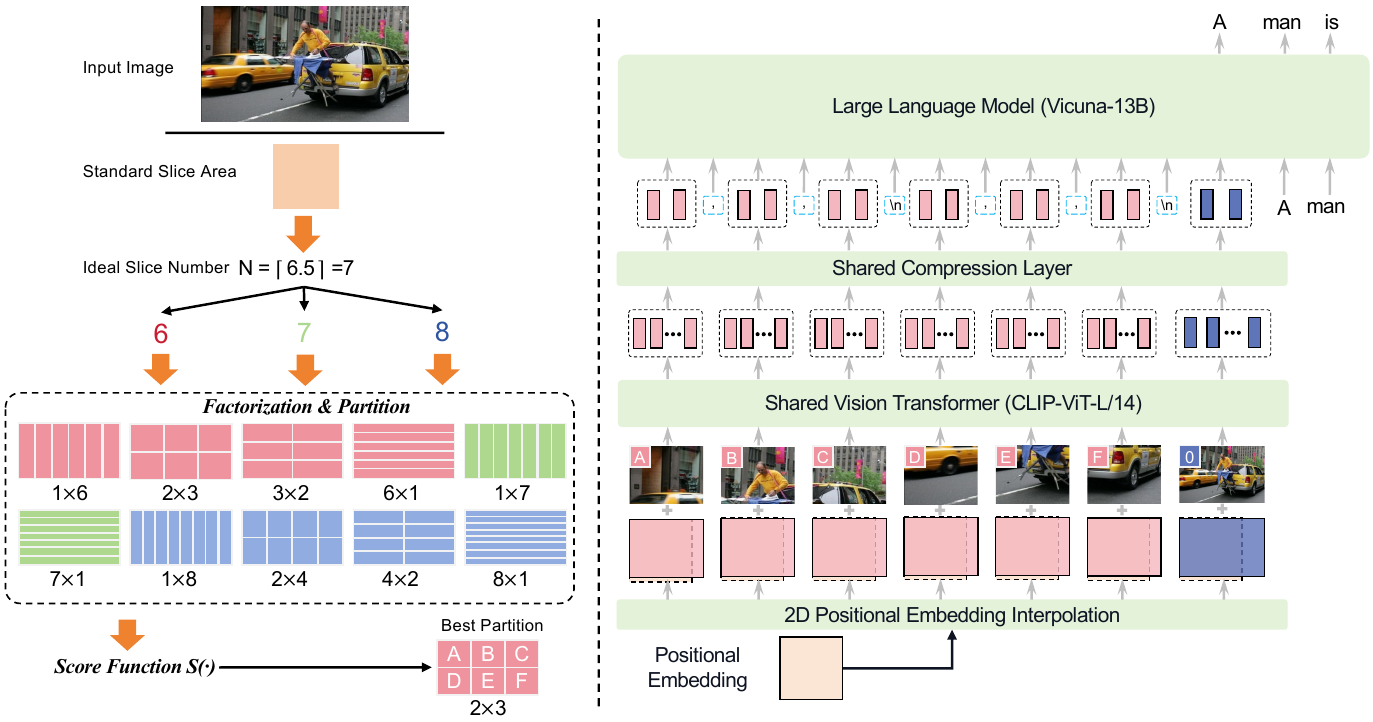}
\caption{The LLaVA-UHD framework. Left: Given a high-resolution image, LLaVA-UHD first calculates the ideal slice number, and then selects the best partition from possible factorization, splitting the high-resolution image into varied-sized slices. Right: Slices are encoded in native aspect ratios by 2D interpolation on position embeddings, and then compressed and arranged in a spatial schema for LLM processing.}
\label{fig:framework}
\end{figure*}


\section{Method}
Based on the principles learned from the pilot experiments, we propose LLaVA-UHD, a large multimodal model that can efficiently perceive any aspect ratio and high-resolution images. As shown in Fig.~\ref{fig:framework}, the model includes three key components: (1) An image modularization strategy that divides native-resolution images into smaller variable-sized slices for efficient and extensible encoding, (2) a compression module that further condenses image tokens from visual encoders, and (3) a spatial decoration schema to organize slice tokens for LLMs.

\subsection{Modularized Visual Encoding}
\label{sec:encoding}

To deal with high-resolution images with varied aspect ratios, a naive approach is to interpolate the position embeddings of ViT to the target shape for direct encoding as a whole. However, this approach is sub-optimal due to the quadratic computation cost and the performance degradation from out-of-distribution issues. To address the challenge, we present a modularized visual encoding strategy. The basic idea is to divide native-resolution images into smaller variable-sized slice slices, where the shape of each slice does not deviate too far from the standard pretraining setting of ViT. With variable-sized slice slices, LLaVA-UHD can achieve full adaptivity to native-resolution images without padding or shape-distorting reshaping.

\smallskip
\textbf{High-Resolution Image Partition Strategy.} The goal of image slicing strategy is to determine a split of high-resolution images, with minimal changes to the resolutions of each slice. Given an image in resolution $(W_I, H_I)$ and a ViT pretrained in resolution $(W_v, H_v)$, we first determine the number of slices (i.e., the ideal computation) needed to process the image: $N=\lceil \frac{W_I\times H_I}{W_v\times H_v} \rceil$. Then we factorize the slice number $N$ into $m$ columns and $n$ rows: $\mathbb{C}_N= \{(m, n)| m\times n = N, m\in \mathbb{N}, n\in \mathbb{N} \}$. To select the most appropriate partition, we define a score function to measure the deviation from the standard pretraining setting of ViT:

\begin{equation}
\small
    S(W_I, H_I, W_v, H_v, m, n)= -\left| \log \frac{W_I \times n}{H_I \times m} - \log \frac{W_v}{H_v}\right|,
\end{equation}
where higher score $S(\cdot)$ indicates a smaller deviation from the standard setting of ViT, and is thus preferred. Therefore the partition can be obtained as follows: 

\begin{equation}
\small
    m^*, n^* = \argmax_{(m,n)\in \bar{\mathbb{C}}} S(W_I, H_I, W_v, H_v, m, n),
\label{equ:partition}
\end{equation}
where the candidate set $\bar{\mathbb{C}} = \mathbb{C_N}$. In practice, we notice that in some cases, there might be only a few possible factorization schemes for $N$, especially for prime numbers, which can lead to limited choices and therefore extreme partitions of images. For example, $N=7$ has only two extreme partition choices, 1:7 and 7:1. To address the issue, in addition to the ideal slice number $N$, we also allow a modest change of slice numbers $N-1, N+1$ to incorporate more plausible partition choices. Therefore, the final partition is given by Equation~\ref{equ:partition}, where $\bar{\mathbb{C}} = \mathbb{C}_{N-1} \cup \mathbb{C}_{N} \cup \mathbb{C}_{N+1}$.

Theoretically, we show that the partition strategy guarantees minor expected changes and modest worst-case changes with respect to standard pretraining resolution $(W_v, H_v)$ for each slice. Specifically, we show that for input images where $N \leq 20$ and aspect ratio in $[1:6, 6:1]$, the aspect ratio of each slice resides within $[1:2, 2:1]$, and the area of each slice resides within $[0.33W_IH_I, 1.5W_IH_I]$. We refer readers to Section~\ref{sec:proofs} for full proof details.


\smallskip
\textbf{Arbitrary Aspect Ratio Slice Encoding.} Most existing LMMs utilize a static resolution for image slice encoding~\citep{bai2023qwen,liu2023llava1.5,instructblip2023}. This essentially prevents full adaptivity to native resolutions, since only several predefined fixed-shape slices are available. Moreover, the static slice resolution inevitably incurs padding or shape-distorting resizing, which hurts the performance, efficiency, and even correctness as discussed in Section~\ref{sec:pilot_exp}. 

To address the problem, we propose to encode image slices in aspect ratios given by the partition strategy as is. Specifically, we proportionally resize the original image following the aspect ratio, such that the number of patches maximally fits within the pretraining budget $M$ (i.e., the number of position embeddings in ViT). Then we reshape the pretrained 1D position embedding sequence of ViT into 2D format $P \in \mathbb{R}^{q\times q \times l}$ following its pretraining setting, where $M=q\times q$, and $l$ is the dimension of position embeddings. After that, we 2D-interpolate $P$ to fit the slice resolution given by the partition strategy for visual encoding. In our experiments, we show that ViT and position embedding parameters can be kept frozen during pretraining, and updating these parameters during the instruction-tuning stage is sufficient for good performance. In addition to slices, we also provide a low-resolution overview image in native aspect ratio. The overview image can provide coarse-grained information and global semantic connections in images.

\subsection{Compression Layer}
High-resolution images require LLMs to process significantly more visual tokens, which accounts for a major part of the computation. For example, a $672\times 1008$ resolution image will produce 3,456 visual tokens for LLaVA-1.5~\citep{liu2023llava1.5}. To address the issue, we compress the visual tokens of each image slice using a shared perceiver resampler layer~\citep{Alayrac2023Flamingo}. Specifically, image tokens output by the visual encoders are resampled to a lower number using a set of query vectors via cross-attention (from $576$ to $64$ in our experiments). Compared with the prevalent MLP-based visual projection approaches~\cite{liu2023llava1.5,2023llava1.6,wang2023cogvlm}, perceiver resampler maintains a fixed and affordable number of visual tokens regardless of image resolutions, and is therefore more compatible with high-resolution image understanding. As a result, LLaVA-UHD can encode $672\times1008$ resolution images using an even lower computation cost than LLaVA-1.5 in encoding $336\times336$ resolution images.

\subsection{Spatial Schema for Image Slices}
Since the image partition is dynamic across different images, it is necessary to inform LLM of the spatial organizations of image slices. Inspired by~\citep{fuyu2023}, we design a spatial schema to inform the relative positions of image slices using two special tokens. Specifically, we use ``,'' to separate the slice representations in a row, and use ``\textbackslash n'' to separate different rows. In our experiments, we find that the simple schema can effectively inform the dynamic partition to yield good performance.

\section{Experiments}
In this section, we empirically investigate the effectiveness of LLaVA-UHD. We first provide the implementation details, and report the evaluation results on 9 common benchmarks compared with strong baselines. Then we provide analytic results for better understanding of the model.

\subsection{Implementation Details}
\textbf{Model Configuration.}
In this work, we built LLaVA-UHD following the implementation of LLaVA-1.5~\citep{liu2023llava1.5}. Specially, we use the CLIP-ViT-L/14\hspace{0.5mm} as visual encoder (default resolution ${336\times336}$), Vicuna-13B~\citep{chiang2023vicuna} as LLM, and a shared visual resampler~\citep{bai2023qwen} as the projector to connect the visual encoder and LLM. During the encoding of image slices, a minor reshape within half patches (maximum 7-8 pixels) could be performed to fit the slice into patches. The number of learnable queries in resampler is set to 64. For the image partitioned as $N$ sub-patches, the number of visual tokens fed into LLM is $64\times(N+1)$, with tokens of the low-resolution overview image. We set the maximum $N$ to be 6 in experiments, which supports a maximum of $672\times1008$ resolution images. Following LLaVA-1.5, we perform a two-stage training as follows. 

\textbf{Stage 1: Pretraining details.}
During this stage, only the perceiver resampler is tuned, with the CC-595K dataset~\citep{liu2024llava} for 1 epoch, using AdamW optimizer with a learning rate of $1e^{-3}$ and the cosine learning rate schedule. The global batch size is set to 256. The training cost of this stage is $\sim$5 hours using 8$\times$A100 GPUs.

\textbf{Stage 2: Instruction-tuning details.}
During this stage, the visual encoder is frozen and we fine-tune the visual resampler and LLM, with a 656K mixture dataset~\citep{liu2023llava1.5} which contains LLaVA-Instruct~\citep{liu2024llava}, TextVQA~\citep{singh2019textqa}, GQA~\citep{hudson2019gqa}, OCR-VQA~\citep{mishra2019ocrvqa}, and Visual Genome~\citep{krishna2017vg}. The learning rate is 2$e^{-5}$ and batch size is 128. Other settings are the same as stage 1. The training cost of this stage is $\sim$18 hours using 8$\times$A100 GPUs.

\subsection{Experimental Setting}
We introduce experimental settings, including the benchmarks, evaluation metrics, and baselines.

\textbf{Benchmarks.} We adopt 9 popular benchmarks to evaluate our model, including: (1) General visual question answering benchmarks such as VQA-V2~\citep{antol2015vqa}, GQA~\citep{hudson2019gqa}, ScienceQA~\citep{lu2022scienceqa}, and VizWiz~\citep{gurari2018vizwiz}; (2) Optical character based visual question answering benchmark such as TextVQA~\citep{singh2019textqa}; (3) Hallucination benchmark such as POPE~\citep{li2023pope}; (4) Comprehensive benchmarks such as MME~\citep{fu2023mme}, MMBench~\citep{liu2023mmbench}, and MMBench-CN~\citep{liu2023mmbench}.

\textbf{Evaluation Metrics.} In addition to the performance on popular benchmarks, we also report the computation cost (TFLOPs) in processing an image in the maximum supported resolution. The computation cost is aggregated from the visual encoder, projector, and LLM. We also report the accumulated multimodal training data volume for reference, which includes image-text pairs used during pertaining and instruction tuning. For models post-trained on existing multimodal models as backbones, this also includes the training data of the backbones.

\textbf{Baselines.} We compare our model with strong baselines. (1) General baselines. We adopt Qwen-VL~\citep{bai2023qwen}, LLaVA-1.5~\citep{liu2023llava1.5}, MiniGPT-v2~\citep{chen2023minigptv2}, Shikra~\citep{chen2023shikra}, BLIP-2~\citep{li2023blip2} and InstructBLIP~\citep{instructblip2023} as representative general baselines. Since the implementation of LLaVA-UHD is highly aligned with LLaVA-1.5, it serves as the most direct baseline. (2) High-resolution LMMs. SPHINX~\citep{SPHINX2023} and mPLUG-Owl2~\citep{ye2023owl2} encode images in fixed resolutions; Ureader~\citep{ye2023ureader} and Monkey~\citep{li2023monkey} support enumerated resolution types (several predefined fixed-shape slices); Fuyu-8B~\citep{fuyu2023} and OtterHD-8B~\citep{li2023otterhd} can encode images in any resolutions.

\begin{table}[tb]
  \caption{Main results on 9 popular benchmarks. \#Data: accumulated multimodal training data volume, MaxRes.: maximum resolution supported, TFLOPs: computation cost of processing maximum resolution images, AR.: aspect ratio supported, ${\rm \Delta}$: improvements over LLaVA-1.5 backbone.}
  \centering
  \resizebox{\columnwidth}{!}{
  \begin{tabular}{@{}cccccccccccccccccccc@{}}
    \toprule
    Model &\#Data & MaxRes. &AR. & TFLOPs   &  VQA$^\mathrm{v2}$ & GQA  & VQA$^\mathrm{T}$ & POPE & SQA & VizWiz & MME & MMB & MMB$^\mathrm{CN}$ \\
    \midrule
    BLIP-2~\cite{li2023blip2} &129M   &224$\times$224 &Fix &1.0    &41.0    &41.0     &42.5  &85.3 &61.0 &19.6 &1293.8  &-   &-      \\
    InstructBLIP~\cite{instructblip2023} &130M  &224$\times$224 &Fix &1.0     &- &49.5  &50.7  &78.9 &63.1 &33.4 &1212.8  & - & -   \\
    Shikra ~\cite{chen2023shikra} &6M   &224$\times$224 &Fix &8.0     &77.4 &-  &-  &-  &-  &-  &-   &58.8   &-   &  \\
    Qwen-VL~\cite{bai2023qwen} &1.4B  &448$\times$448 & Fix &9.2     &78.8 &59.3  &\underline{63.8}  &- &67.1 &35.2  &-   &38.2  &7.4     \\
    SPHINX~\cite{SPHINX2023} &1.0B  &448$\times$448 &Fix &39.7      &78.1   &62.6       &51.6 &  80.7 &69.3 &39.9 & 1476.1  &66.9   &56.2      \\
    SPHINX-2k~\cite{SPHINX2023} &1.0B  &762$\times$762 &Fix &69.4      &\underline{80.7}   &63.1    &61.2 &\underline{87.2} &70.6 &44.9 &1470.7  &65.9   &57.9    \\
    MiniGPT-v2~\cite{chen2023minigptv2} &326M &448$\times$448  &Fix &4.3      &-   &60.1    &-   &- &- &53.6    &-         &-   &-       \\
    Fuyu-8B~\cite{fuyu2023} &- &1024$\times$1024 &Any &21.3       &74.2   &-        &- &74.1 &- &-   &728.6 &10.7   &-        \\
    OtterHD-8B~\cite{li2023otterhd} &-  &1024$\times$1024 &Any &21.3       &-        &-   &- &86.0 &- &-   &1223.4   &58.3   &-           \\
    mPLUG-Owl2~\cite{ye2023owl2} &401M  &448$\times$448  &Fix  &1.7     &79.4   &56.1  &58.2   &86.2  &68.7  &54.5 &1450.2 &64.5   &-         \\
    UReader~\cite{ye2023ureader} &86M  &896$\times$1120 &Enum &26.0     &-  &-     &57.6 &- &- &-  &-         &-        &-     \\
    Monkey~\cite{li2023monkey} &1.0B  &896$\times$1344 & Enum &65.3       &80.3   &60.7      &- &67.6  &69.4 &\textbf{61.2}   &-         &-        &-     \\
    LLaVA-1.5~\cite{liu2023llava1.5} &1.2M & 336$\times$336 &Fix & 15.5    &  80.0 & \underline{63.3}  & 61.3 & 85.9 & \underline{71.6} & 53.6 & \underline{1531.3} & \underline{67.7} & \underline{63.6}  \\
    \midrule
    LLaVA-UHD (ours) &1.2M& 672$\times$1008 & Any & 14.6      &\textbf{81.7}  &\textbf{65.2}   & \textbf{67.7} & \textbf{89.1} &  \textbf{72.0} & \underline{56.1} & \textbf{1535.0} & \textbf{68.0} & \textbf{64.8}  \\
    ${\rm \Delta}$ &- &{\color{black} $\times$6 times}   &- &{\color{black}-0.9} &{\color{black} +1.7} &{\color{black} +1.9}  &{\color{black} +6.4} &{\color{black} +3.2} &{\color{black} +0.4} &{\color{black} +2.5} &{\color{black} +3.7} &{\color{black} +0.3} &{\color{black} +1.2}\\
  \bottomrule  \end{tabular}
  }
\label{tab:sota}
\end{table}


\subsection{Main Results}

We report the main experimental results in Table~\ref{tab:sota}, from which we have the following observations: (1) LLaVA-UHD outperforms strong baselines on popular benchmarks. This includes strong general baselines trained on 2-3 orders of magnitude more data such as Qwen-VL and InstructBLIP, and also high-resolution LMMs that require significantly more computation such as Fuyu-8B, OtterHD-8B, Monkey and SPHINX-2k. The results show that LLaVA-UHD can properly deal with native-resolution images for strong performance, as well as good data and computation efficiency. (2) LLaVA-UHD achieves significant improvements over the LLaVA-1.5 backbone. Notably, by simply perceiving images in native high-resolution, LLaVA-UHD achieves 6.4 accuracy improvement on TextVQA and 3.2 accuracy improvement on POPE. The reason is that the blurred content in low-resolution images can prevent LMMs from accurately identifying the challenging fine-grained objects and optical characters. The results demonstrate the fundamental role of perceiving native high-resolution images in various multimodal tasks, and the effectiveness of LLaVA-UHD in addressing the problem. (3) In terms of resolution and efficiency, compared with LLaVA-1.5 associated fixed $336\times336$ resolution, LLaVA-UHD supports 672$\times$1088 resolution images in any aspect ratio using only 94\% inference computation. The results indicate promising scalability of LLaVA-UHD to potentially larger resolutions in future.

\begin{table}[b]
  \caption{Ablation Results. FP: Fixed image partition strategy.
  }
  \label{tab:module_ablations}
  \centering
  \small
  \resizebox{0.8\columnwidth}{!}{
  \begin{tabular}{@{}lccccccccccccccccccccc@{}}
    \toprule
    Model & \#TFLOPs    &  VQA$^\mathrm{v2}$ & GQA  & VQA$^\mathrm{T}$  & POPE & SQA & VizWiz \\
    \midrule
    LLaVA-1.5   & 15.50   &  80.0 & 63.3 & 61.3 & 85.9 & 71.6 & 53.6 \\
    \hspace{0.8em} w/ adaptive enc.    & 15.50      &  \textbf{80.1} & \textbf{64.1} & \textbf{61.8}  & \textbf{86.7} & \textbf{72.0} & \textbf{54.2} \\
    \midrule
        LLaVA-UHD  & \textbf{14.63}      &\textbf{81.7}  &65.2   & \textbf{67.7} &89.1 & \textbf{72.0} & 56.1   \\
        \hspace{0.8em} w/ MLP & 113.65    &  81.6  &\textbf{65.4}  & 66.9 & \textbf{89.2} &  71.5 & \textbf{56.3} \\
    \hspace{0.8em} w/ MLP \& FP.~\citep{SPHINX2023} & 80.10    &  81.2  & 64.3    & 66.1  & 89.1   & 71.1  & 54.3 \\
    \hspace{0.8em} w/o spatial schema & 80.07 &  79.3  &  63.9  & 65.4  &  88.3  & 70.3  & 53.1 \\


  \bottomrule 
  \end{tabular}
  }
\end{table}

\smallskip
\subsection{Analytic Results}
We provide further analytic results, including ablation on alternative components, evaluation on images with more extreme aspect ratios, best practice for frozen/trainable parameters, and case study.

\smallskip
\textbf{Ablation Study.} In Table~\ref{tab:module_ablations}, we conduct ablation studies on alternative components. (1) We replace the padding strategy of LLaVA-1.5 with the adaptive encoding strategy of LLaVA-UHD, supporting arbitrary aspect ratios while maintaining identical maximum resolutions. We can observe consistent improvement since wasted computation from padding is avoided. (2) We replace the perceiver resampler of LLaVA-UHD with the 2-layer MLP of LLaVA-1.5. We observe that perceiver resampler achieves comparable or better performance than MLP, using only 12.9\% computation cost. (3) We further replace the LLaVA-UHD image partition strategy with the naive partition strategy~\citep{SPHINX2023} (i.e., fixed $2\times2$ slices). Results show that LLaVA-UHD can more properly divide images into slices for better performance. (4) We remove the spatial schema from LLaVA-UHD. The performance degradation demonstrates the effectiveness and necessity of spatial schema in informing the dynamic slice positions for LMMs.

\smallskip
\textbf{LLaVA-UHD generalizes to images with extreme aspect ratios.} We further investigate the generalization capability of LLaVA-UHD by constructing an extended version of existing benchmarks. Specifically, we expand the aspect ratio of an image by doubling the length of its longer side through padding. From the results in Table~\ref{tab:padding_evaluation}, we can see that the advantage of LLaVA-UHD increases as compared with LLaVA-1.5 and alternatives. The reason is that LLaVA-UHD perceives images in native aspect ratios. In comparison, LMMs that encode images in fixed aspect ratios will suffer from significant distortion in the content shapes. Moreover, this also causes the computation to be unevenly distributed along the width and height of the image content.

\begin{table}[tb]
  \caption{Experimental results on extreme aspect ratio images. Absolute performance and the degradation compared with the standard benchmarks in Table~\ref{tab:module_ablations} are reported.
  }
  \label{tab:padding_evaluation}
  \centering
  \resizebox{\columnwidth}{!}{
      \begin{tabular}{@{}lcccccccccccc@{}}
        \toprule
        Model   & \#TFLOPs  & VQA$^{\mathrm{v2}}$ & GQA & VQA${^\mathrm{T}}$ & POPE & SQA & VizWiz \\
        \midrule
        LLaVA-1.5   & 15.50 & 74.6 (-5.4) & 57.9 (-5.4) & 58.4 (-3.9) & 81.1 (-4.8)& 66.3 (-5.3)& 50.1  (-3.5)\\
        \hspace{0.8em} w/ adaptive enc.   & 15.50    & \textbf{74.9} (-5.2) & \textbf{62.5} (-1.6) & \textbf{60.7} (-1.1) & \textbf{82.3}  (-4.4) & \textbf{66.9} (-5.1) & \textbf{50.9} (-3.3) \\
        \midrule
        LLaVA-UHD  & 14.63 & \textbf{81.4} (-0.3)  &  61.8 (-3.4) & \textbf{64.5} (-3.2) & 85.1 (-4.0)  &  \textbf{71.5} (-0.5) &  54.0 (-2.1)\\
            \hspace{0.8em} w/ MLP & 113.65 & 81.3 (-0.3)  &  \textbf{62.0} (-3.4) & 63.9 (-3.0) & \textbf{85.2} (-4.0) &  70.9 (-0.6) & \textbf{54.3} (-2.0)  \\
        \hspace{0.8em} w/ MLP \& FP.~\citep{SPHINX2023} & 80.10   & 79.6 (-1.6) & 61.9 (-2.4)  & 58.5 (-7.6)  &84.4 (-4.7) & 69.4 (-1.7) & 52.2 (-2.1)     \\
      \bottomrule  \end{tabular}
  }
\end{table}

\begin{table}[t]
  \caption{The effect of tuning visual encoder at different training stages. 
  }
  \label{tab:frozen_stage}
  \centering
  \small
  \begin{tabular}{@{}cc|ccccccccccccc@{}}
    \toprule
    \multicolumn{2}{c|}{Update ViT} &\multirow{2}{*}{VQA$^\mathrm{v2}$} &\multirow{2}{*}{GQA}  &\multirow{2}{*}{VQA$^\mathrm{T}$} &\multirow{2}{*}{POPE} &\multirow{2}{*}{SQA}  &\multirow{2}{*}{VizWiz} \\
    \cline{1-2}
    pre-training &Fine-tuning \\
    \midrule
    
    & \checkmark          & \textbf{81.7}  & \textbf{65.2} & \textbf{67.7} & \textbf{89.1}  &  \textbf{72.0} &  \textbf{56.1}   \\
    \checkmark &          & 78.2  &  61.1 & 58.9 & 83.9 & 68.6 &  51.4  \\
    &                     & 79.4  &  64.5 & 65.7 & 87.3 & 71.9 &  55.4   \\
    \checkmark &\checkmark &80.2  &  63.7 & 62.6 & 87.2 & 71.6 &  55.1   \\
    \midrule
    \multicolumn{2}{c|}{LLaVA-1.5~\citep{liu2023llava1.5}} &  80.0 & 63.3 & 61.3 & 85.9 & 71.6 & 53.6 \\
  \bottomrule  \end{tabular}
\end{table}

\begin{figure*}[t]
\centering
\includegraphics[width=1.0\linewidth]{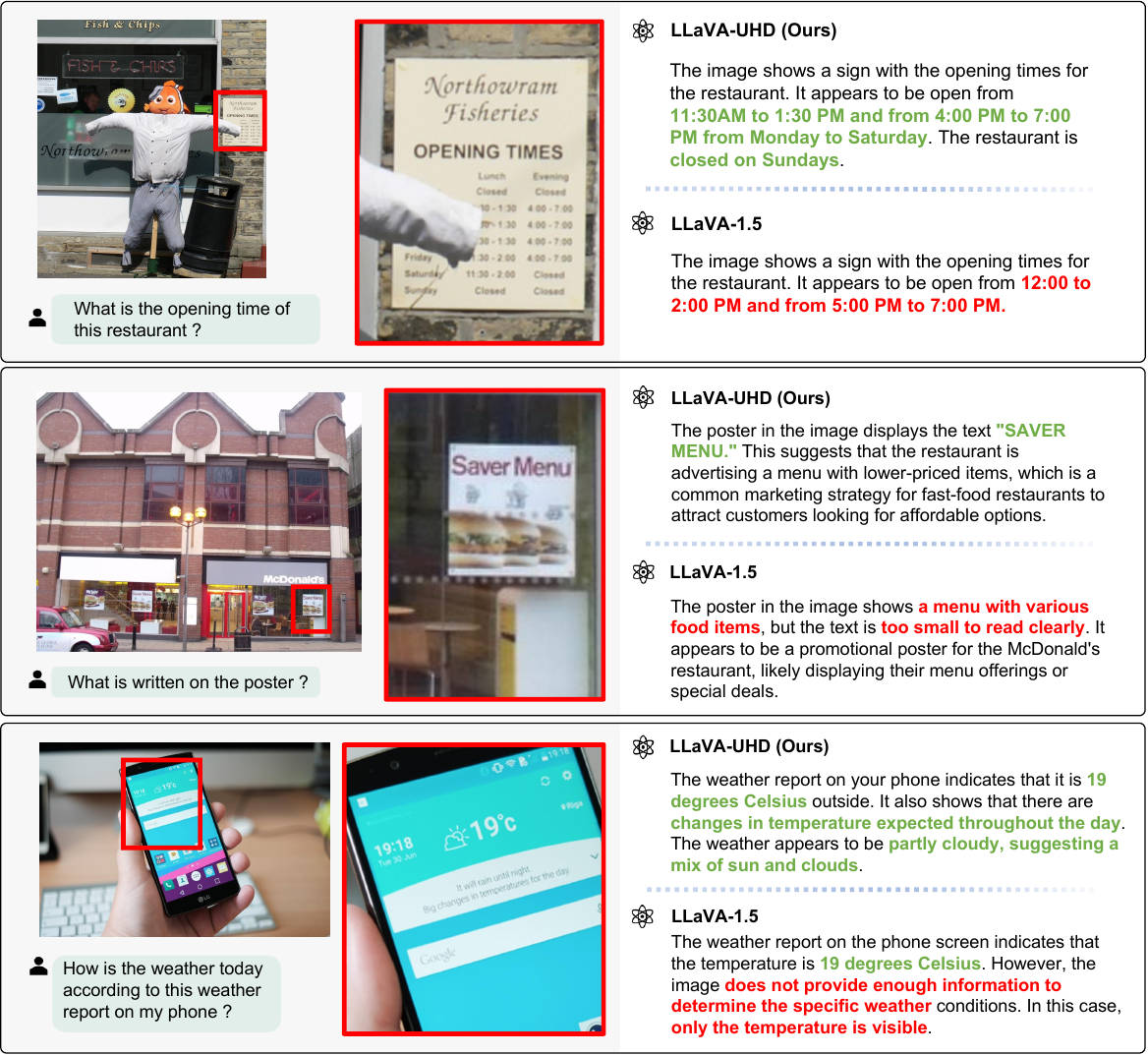}
\caption{Qualitative comparison of LLaVA-UHD and LLaVA-1.5 in fine-grained recognition and reasoning capabilities.}
\label{fig:case}
\end{figure*}

\smallskip
\textbf{Instruction-tuning ViT parameters is sufficient for adaptation.} We investigate the effect of tuning ViT parameters at different training stages, including pretraining and instruction-tuning. From the results in Table~\ref{tab:frozen_stage}, we observe that: (1) Updating ViT during instruction-tuning is sufficient to achieve good performance. In fact, we find that LLaVA-UHD can improve over LLaVA-1.5 even when ViT parameters are frozen in both pretraining and instruction tuning. (2) Further updating ViT during pretraining does not lead to better results. We hypothesize the reason is that jointly training ViT and resampler (from scratch) on limited pretraining data can lead to instability issues.

\smallskip
\textbf{Case Study.} To provide a more intuitive understanding of the capabilities of LMMs in dealing with high-resolution images, we provide qualitative results for LLaVA-UHD and LLaVA-1.5 in Fig.~\ref{fig:case}. We can see that LLaVA-UHD can correctly identify the dense content in the timetable (Case 1), the text on the small poster (Case 2), and icons and text on the phone (Case 3) for fine-grained recognition and reasoning. In comparison, LLaVA-1.5 can only perceive coarse-grained information, and therefore tends to provide either uninformative (Cases 1 and 2) or incorrect/hallucinated answers (Case 3) in these challenging scenarios. The results demonstrate the effectiveness and advantage of LLaVA-UHD in perceiving native aspect ratio and high-resolution images for fine-grained multimodal capabilities.

\section{Related Work}

\textbf{Visual Encoding in LMMs.}
The advent of ChatGPT~\citep{ChatGPT2022} and GPT-4~\citep{achiam2023gpt4} has spurred the development of numerous open-source large language models (LLMs)~\citep{chiang2023vicuna,touvron2023llama,Chung2022Flan5}. Utilizing an LLM as a language encoder and decoder, there springs up plenty of LMMs~\citep{li2023blip2,instructblip2023,Alayrac2023Flamingo,liu2024llava,bai2023qwen,hong2023cogagent}, with aim at understanding visual image. Therefore, how to encode vision features into LLMs becomes the core problem in the community. Fortunately, CLIP~\citep{radford2021clip} proposes to respectively extract language embeddings using language models like BERT~\citep{Devlin2018BERT} and visual features using vision models like ViT~\citep{dosovitskiy2020vit} and CNN~\citep{he2016resnet}, and align them in contrastive learning fashion using considerable image-text pairs~\citep{schuhmann2022laion}, so that visual embeddings are well aligned towards the language. 

Existing visual projection approaches towards LLMs can be divided into three categories. (1) Flamingo~\citep{Alayrac2023Flamingo} proposes perceiver resampler, which utilizes a fixed number of queries to capture visual features by cross-attention operation and feeds them into LLMs for image/video understanding. 
(2) BLIP-2~\citep{li2023blip2} pretrains Q-Former to bridge the image encoder and LLMs. (3) LLaVA~\citep{liu2024llava} just leverages an MLP module to connect language and vision feature space. Beyond them, SPHINX~\citep{SPHINX2023} mixes many kinds of visual features, including DINO-V2~\citep{oquab2023dinov2}, CLIP-ViT~\citep{radford2021clip} and CLIP-CNN~\citep{radford2021clip}, and Q-Former to augment visual representation. Vary~\citep{wei2023vary} pretrains a visual model tailored for document/chart recognition and understanding, and integrates it with visual features of LLaVA~\citep{liu2024llava} for further feature enhancement. 

However, since these LMMs rely on CLIP-ViT that requires fixed resolution image as input, it hinders LMMs from handling images with higher resolution or any aspect ratio, and undermines fine-grained downstream tasks like optical character recognition or small object understanding.

\smallskip
\textbf{High-resolution LMMs.}
To perceive images with higher resolutions, recent work can be divided into four categories. (1) Up-Resize. Qwen-VL~\citep{bai2023qwen} interpolates the positional embedding of ViT from 224$\times$224 to 448$\times$448 and additionally executes a training stage to fine-tune the ViT. CogAgent~\citep{hong2023cogagent} and LLaVA-HR~\citep{luo2024feast} marries a large low-resolution encoder with a small high-resolution image. MiniGPT-v2~\citep{chen2023minigptv2} only resizes the positional embeddings without fine-tuning the visual encoder during instruction tuning. These methods dramatically change the original visual position encoding of CLIP-ViT~\citep{radford2021clip}, which can cause sub-optimal visual representation.
(2) Fix+Crop. To address the above issue, SPHINX~\citep{SPHINX2023} utilizes a fixed window size (224$\times$224) to crop a padded image (448$\times$448) into four slices, and concatenates them with a down-sampled 224$\times$224 image as visual inputs. Monkey~\citep{li2023monkey} follows this idea yet increases the accessible image size to 896$\times$1344, and converts each slice using a shared resampler. 
(3) Fix+Enumerated-Crop. UReader~\citep{ye2023ureader}, LLaVA-1.6~\citep{2023llava1.6} and infiMM-HD~\citep{liu2024infimm} enumerate a similar aspect ratio to resize, rather than using a fixed square ratio (e.g., 2$\times$2 as in SPHINX~\citep{SPHINX2023}). 
The unavoidable image resizing and padding operation might cause image deformation and waste of computation, respectively.
(4) Any. Fuyu-8B~\citep{fuyu2023} and Otter-HD~\citep{li2023otterhd} directly utilize LLMs to encode visual features instead of vision transformers. They just split images into patches and project them using linear layers before feeding into the LLM. Regarding image patches as a sequence enables itself to process images with continuous resolution.
However, the removal of an image encoder means insufficient visual representation, which makes these methods limited in unsatisfactory performance. 

In comparison, LLaVA-UHD supports images in any aspect ratios and high resolutions. By integrating the advantages of modularized and adaptive image encoding, as well as perceiver resampler, LLaVA-UHD can achieve strong performance with improved computation efficiency.

\section{Conclusion}
In this work, we present LLaVA-UHD, a large multimodal model that efficiently perceives any aspect ratio and high-resolution images. Comprehensive experimental results on 9 popular benchmarks demonstrate the effectiveness of LLaVA-UHD, especially in fine-grained multimodal capabilities. Analytical evaluation results are provided for deeper understanding of the model. In this work, we limit the resolution of LLaVA-UHD to maximum $672\times1008$. In future, considering the promising efficiency and scalability, we will explore higher-resolution images and more challenging tasks such as small object detection and segmentation. Besides, image slices are currently independently encoded, with interactions only in LLMs. We plan to establish efficient connections between image slices via improved visual encoding strategies for fine-grained global information interaction.


{
\small
\bibliographystyle{plainnat}
\bibliography{references}
}

\newpage

\appendix

\section{Detailed Illustration on GPT-4V Phases}
\label{sec:GPT-4V-illustration}

\begin{figure*}[t]
\centering
\includegraphics[width=1.0\linewidth]{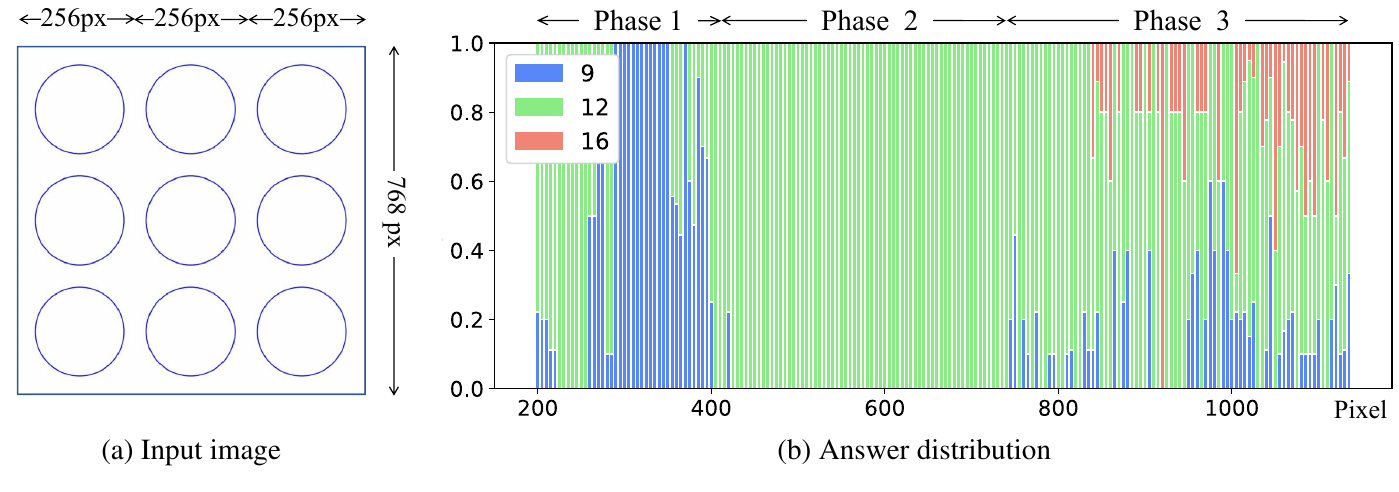}
\caption{Results on probing GPT-4V via continuously changing image resolutions.}
\label{fig:gpt4v_exp2_appendix}
\end{figure*}

\begin{figure*}[t]
\centering
\includegraphics[width=0.85\linewidth]{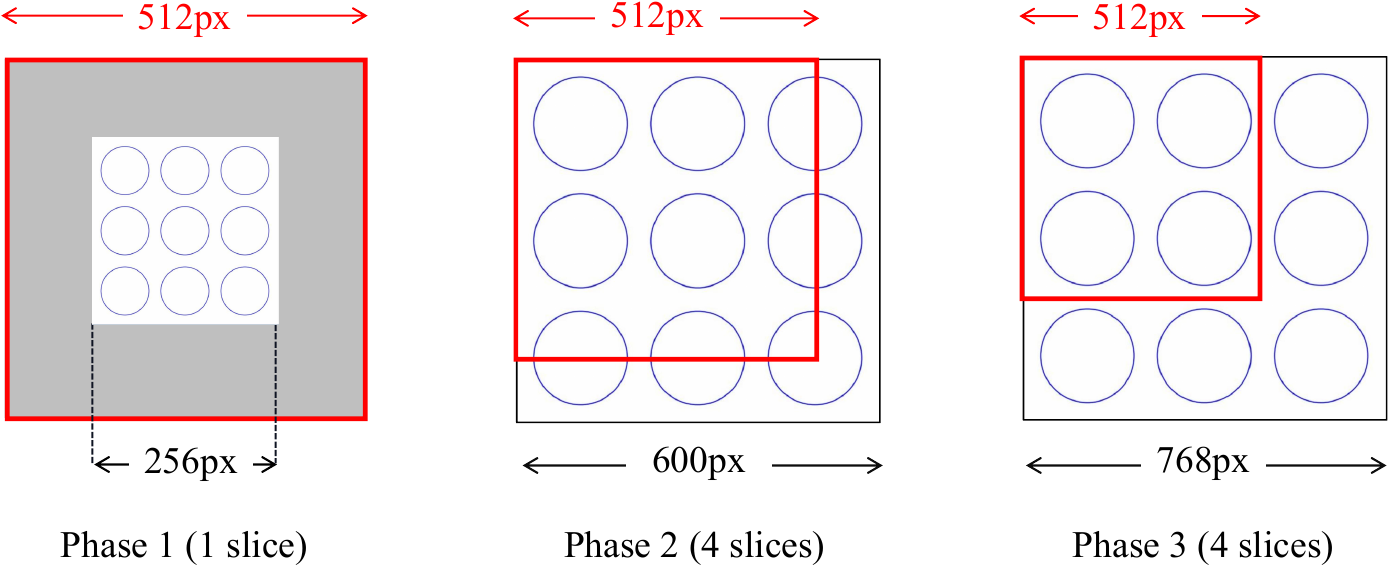}
\caption{ Illustration on GPT-4V phases (hypothesized). Red square indicates a slice.}
\label{fig:illustration}
\end{figure*}

 From the pilot experimental results in Fig.~\ref{fig:gpt4v_exp2_appendix}, we observe that the GPT-4V responses show a significant phase change with image resolutions. Here we provide detailed illustrations of the hypothesized cause from the perspective of visual encoding:

(1) In phase 1, since there is only one image slice, most answers are correct. More specifically, when dealing with input images under 512 resolution, if the images are resized to 512, the behavior will be the same within phase 1. However, since the behavior changes significantly within phase 1, we suspect that the input images are most likely to be padded into 512 resolutions, as shown in Fig.~\ref{fig:illustration}(a).

(2) In phase 2, answer 12 dominates the responses possibly due to the incomplete circles in each slice, as shown in Fig.~\ref{fig:illustration}(b).

(3) Phase 3 shows mixed answers of 9, 12 and 16. Among these responses, answer 16 can be well explained by the slice strategy in Fig.~\ref{fig:illustration}(c). Besides, we also notice that many abnormal phenomenons in Fig.~\ref{fig:gpt4v_exp2}(b) cannot be perfectly explained yet, which we leave for future work.

\section{Proofs}
\label{sec:proofs}
In this section, we provide proofs for the image partition strategy. We show that the slice resolution exhibits modest changes to the original resolution of ViT.

\smallskip
\textbf{Range of Slice Aspect Ratios.} The aspect ratio of the slice can be represented by: 
\[
\frac{W_v}{H_v} = \frac{W_I}{m} : \frac{H_I}{n},
\]
where $W_v$, $H_v$ are the width and height of the slice, $W_I$, $H_I$ are the sizes of the original image, and (m, n) is the best partition. Restricting  the aspect ratio $r = \frac{W_v}{H_v} \in [\frac{1}{2} , 2] $ is equivalent to $\left|\log(\text{r})\right| \leq \left| \log 2 \right|$, which is also equivalent to $\left| \log\left(\frac{W_I}{H_I}\right) - \log(\frac{n}{m}) \right| \leq \left| \log(2) \right|$. We need to prove: 
\[
\forall \frac{W_{I}}{H_{I}} \in [\frac{1}{6}, 6], N \leq 20 
\]
\[
\exists (\mbox{m, n}) \in \bar{\mathbb{C}}, \left| \log\left(\frac{W_{I}}{H_{I}}\right) - \log(\frac{n}{m}) \right| \leq |\log(2)|,
\]
which is equivalent to
\[
 \forall N \leq 20, (n_{i}, m_{i}) \in \bar{\mathbb{C}} 
\]
\[
\exists (n_{j}, m_{j}) \in \bar{\mathbb{C}}, \left| \left(\log\left(\frac{n_{i}}{m_{i}}\right) - \log\left(\frac{n_{j}}{m_{j}}\right) \right) \right| \leq 2 \cdot \left|\log(2)\right|,
\]
which can be verified by enumerating all possible factorizations of $\bar{\mathbb{C}} =  \mathbb{C}_{N-1} \cup \mathbb{C}_{N} \cup \mathbb{C}_{N+1}$ for $N \leq 20$. The results show that the aspect ratio of each slice resides within $[\frac{1}{2}, 2]$. 

\smallskip
\textbf{Expected Aspect Ratio.} We assume that the ratio of the original image is greater than 1 (i.e., $H_I > W_I$). The situation is the same for $H_I < W_I$. Assuming that the sizes of the images are uniformly distributed for $N\in [0, 20]$, while the aspect ratio of the original images $\frac{W_I}{H_I} \in [1, 6]$, we have $P(W_I,W_H,n,m) = \frac{1}{20} \cdot \frac{1}{5}$. The expected aspect ratio can be obtained by:

\[
\small
\E(\frac{m \times W_I}{n \times H_I}) = \iint_{{\begin{array}{c}
    \frac{W_I}{H_I} \in [1, 6] \\
    W_I \cdot H_I \in [0, 20s] \\
    n,m = \argmax S(\cdot) 
  \end{array}}} (\frac{m \times W_I}{n \times H_I}) \cdot P(W_I,H_I,n,m) \ dW_I dH_I,
\]
where $s$ is the area of a standard resolution of ViT. After calculation, we obtain $\E(r) = 1.258$, $\Var(r) = 0.048$. The results show that the expected aspect ratio of the slices is 1:1.258, which is close to the standard pertaining setting of ViT.
More commonly assuming that images are uniformly distributed between $[1, 3]$, and the aspect ratio is uniformly distributed between $[1, 2]$, we have $\E(r) = 1.147$, $\Var(r) = 0.011$, indicating even smaller changes.

\smallskip
\textbf{Range of Slice Area.} Let $n = \frac{W_I}{W_v} \times \frac{H_I}{H_v}$, which leads to $N= \lceil n \rceil$. We consider dividing the image into $\{N-1, N, N+1\}$ slices. Therefore, the maximum value of each slice $\text{S}_\text{max} = \frac{n}{N-1}$ (when $N \neq 2$), and $\text{S}_\text{max} = \frac{n}{N}$ (when $N = 2$). The minimum value $\text{S}_\text{min} = \frac{n}{N+1}$. As $n$ approaches $3^-$, where $N = 3$, $\text{S}_\text{max}$ achieves the maximum value of $1.5$. Similarly, as $n$ approaches $1^+$, where $N = 2$, $\text{S}_\text{min}$ achieves the minimum value of $0.33$.

\smallskip
\textbf{Expected Slice Area.} Still assuming that the sizes of the images are uniformly distributed within $N \in [0, 20]$, while the aspect ratio of the images $\frac{W_{I}}{H_{I}} \in [\frac{1}{6}, 6]$. The expected area of slice can be obtained by:
\[
\E(\frac{W_I \times H_I}{n \times m}) = \iint_{{\begin{array}{c}
    \frac{W_I}{H_I} \in [1, 6] \\
    W_I \cdot H_I \in [0, 20s] \\
    n,m = \argmax S(\cdot) 
  \end{array}}} (\frac{W_I \times H_I}{n \times m}) \cdot P(W_I,H_I,n,m) d W_I d H_I.
\]
After calculation, we obtain $\E(\frac{W_I \times H_I}{n \times m})= 1.057$, $\Var(\frac{W_I \times H_I}{n \times m})= 0.016$. This shows that our slice areas are relatively concentrated, similar to the original resolution of ViT.

\section{Discussions}
We provide discussions on limitations and potential negative impact of this work.

\smallskip
\textbf{Limitations and Future Work.} (1) Higher resolutions. In this work, we limit the resolution of LLaVA-UHD to maximum $672\times1008$. Although this resolution increases the standard LLaVA-1.5 resolution by 6 times, higher-resolution images such as 4K images and remote sensing images are still out of reach. In future, considering the promising efficiency and scalability, we will explore higher-resolution images and more challenging tasks such as small object detection and segmentation. (2) Joint slice encoding. Currently image slices are currently independently encoded, with interactions only in LLMs. We plan to establish efficient connections between image slices via improved visual encoding strategies for fine-grained global information interaction.

\smallskip
\textbf{Potential Negative Impact.} In this work, we investigate the failure pattern and the underlying cause for GPT-4V and LLaVA-1.5. The mechanism can be potentially used for adversarial attacks on these models. It is worth noting that the goal of this work is to raise attention to the vulnerability of LMMs and provide a deeper understanding of the importance of visual encoding strategies. This work calls for further efforts to mitigate the revealed issues to ensure the robustness and safety of LMMs.

\end{document}